%% file: root.tex
\documentclass[letter, 10pt, conference]{ieeeconf}
\IEEEoverridecommandlockouts
\usepackage{graphicx}
\usepackage{textcomp}
\usepackage{gensymb}
\usepackage{tikz}
\def\checkmark{\tikz\fill[scale=0.4](0,.35) -- (.25,0) -- (1,.7) -- (.25,.15) -- cycle;} 
\usepackage{graphics}
\usepackage{amsmath}
\usepackage{amssymb}
\usepackage{booktabs}

\title{\LARGE \bf
Domain Adaptation for Outdoor Robot Traversability Estimation \\ from RGB data with  Safety-Preserving Loss}

\author{Simone Palazzo$^{1}$, Dario C. Guastella $^{1}$, Luciano Cantelli$^{1}$, Paolo Spadaro$^{1}$, Francesco Rundo$^{2}$\\ Giovanni Muscato$^{1}$,  Daniela Giordano$^{1}$, Concetto Spampinato$^{1}$
\thanks{$^{1}$ Department of Electrical, Electronics and Computer Engineering, University of Catania, 95125 Catania, Italy
        {\tt\small www.dieei.unict.it}}%
\thanks{$^{2}$ ADG Central R\&D, STMicroelectronics, 95121 Catania, Italy
        {\tt\small francesco.rundo@st.com}}%
}

\begin{document}

\maketitle
\thispagestyle{empty}
\pagestyle{empty}

\begin{abstract}
Being able to estimate the traversability of the area surrounding a mobile robot is a fundamental task in the design of a navigation algorithm. However, the task is often complex, since it requires evaluating  distances from obstacles, type and slope of terrain, and dealing with non-obvious discontinuities in detected distances due to perspective. In this paper, we present an approach based on deep learning to estimate and anticipate the traversing score of different routes in the field of view of an on-board RGB camera. 
The backbone of the proposed model is based on a state-of-the-art deep  segmentation model, which is fine-tuned on the task of predicting route traversability. We then enhance the model's capabilities by a) addressing domain shifts through gradient-reversal unsupervised adaptation, and b) accounting for the specific safety requirements of a mobile robot, by encouraging the model to err on the safe side, i.e., penalizing errors that would cause collisions with obstacles more than those that would cause the robot to stop in advance. 
Experimental results show that our approach is able to satisfactorily identify traversable areas and to generalize to unseen locations.
\end{abstract}

\section{INTRODUCTION}
\label{sec:intro}
\input{intro.tex}

\section{RELATED WORKS}
\label{sec:related}
\input{related.tex}

\section{METHOD}
\label{sec:method}
\input{method.tex}

\section{EXPERIMENTAL RESULTS}
\label{sec:exp}
\input{exp_intro.tex}

\subsection{DATA ACQUISITION}
\label{sec:dataset}
\input{dataset.tex}

\subsection{TRAINING DETAILS}
\label{sec:training}
\input{training.tex}

\subsection{PERFORMANCE ANALYSIS}
\label{sec:performance}
\input{performance.tex}

\section{CONCLUSIONS}
\label{sec:conclusions}
\input{conclusions.tex}

\bibliographystyle{IEEEtran}
\bibliography{IEEEfull.bib}
\end{document}

%% file: intro.tex
Traversability analysis is essential for long-term robot operations in real scenarios, especially in outdoor and unstructured environments, since establishing whether spaces will be traversable or not may support better planning and navigation. Our approach aims at performing \textit{traversability anticipation}, which is significantly different --- and more useful in certain operating contexts ---  than classical obstacle detection and avoidance, as it pertains the capability of a robot to identify the traversable horizon in front of the vehicle. Indeed, service robots~\cite{6696421,2832747.2832901}, electric wheelchairs~\cite{5675357} and autonomous cars are expected to take actions that do not affect their safety (and their passengers') in the immediate nor in the future (e.g., following a route ending with a sharp edge). Traversability anticipation is also relevant in other application scenarios where a viable path cannot be trivially identified, such as search and rescue for disaster response~\cite{6719358}, agricultural robotics~\cite{narvaez2018terrain} and planetary exploration~\cite{doi:10.1002/rob.21833}.

\begin{figure}[ht!]
    \centering
    \includegraphics[width=0.49\textwidth]{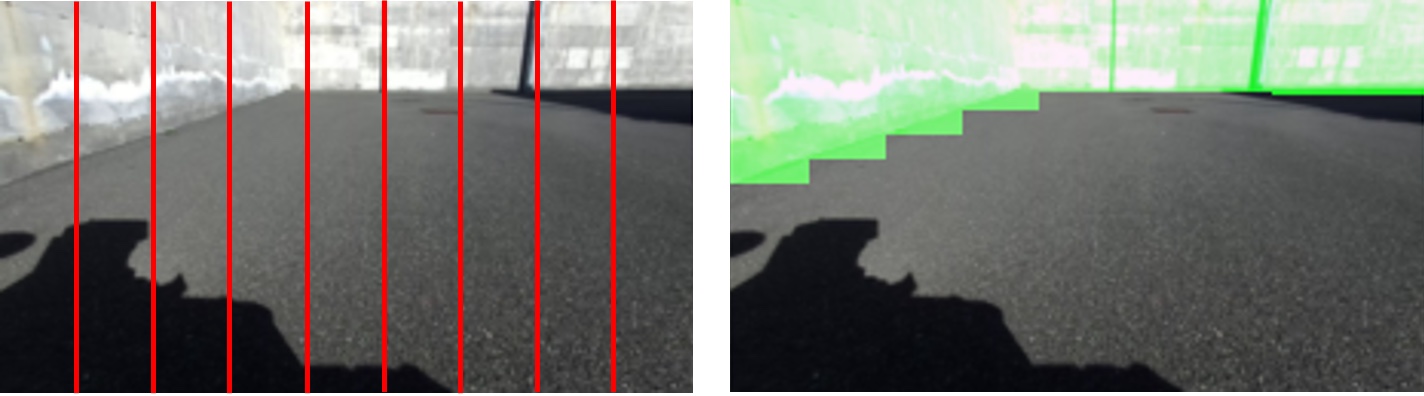}
    \caption{Traversable routes. We split the input images into $k$ bins (in our implementation, $k =$ 9) and then manually provide, for each bin, the traversability score, i.e., a value indicating how far (in terms of image height) a robot can move before running into an obstacle. A value of 1 indicates that the bin is fully traversable, while 0 indicates that it is not. Our model, for each image, predicts (through regression) a $k$-valued vector estimating the traversability (green-colored bins are non-traversable parts) of each bin.}
    \label{fig:traversable}
\end{figure}

Traditional approaches to identify non-traversable paths, e.g., through depth measurements by 2D/3D lidars~\cite{7989182,7139749}, depth cameras~\cite{6698836,balta2013terrain}, stereo camera pairs~\cite{ doi:10.1002/rob.21833,973332}, cannot be employed for traversability anticipation because of the technological limitations that hinder the possibility to obtain reliable measurements at considerable distances (e.g., lidars and depth cameras are sensitive to reflective surfaces and transparent objects, while stereo cameras can fail in depth estimation with poorly-matching images). 
All the mentioned limitations have pushed the research community to operate with RGB data, as in~\cite{4399610,10.5555/647288.721755}, for traversable/non-traversable terrain classification, or in~\cite{HiroseSVGS18}, where fish-eye RGB images are employed for traversability estimation. 
In particular, this last approach proposes a semi-supervised approach, posed as an anomaly detection task, that leverages the recent success of generative adversarial networks (GANs) for generating scenes without obstacles; acquired frames are then compared to the synthesized ones to identify non-traversable areas.
Although this approach yields high accuracy in detecting traversable spaces, we argue that its main limitations are: a) it can detect close obstacles only, thus too late to adequately support action planning; and b) GANs are known to match data distributions, making the approach scarcely generalizable to new domains without requiring a new training from scratch.

In this paper we, instead, propose a deep model for anticipating traversable routes, coping, at the same time, with domain shifts in an unsupervised manner. More specifically, our method leverages the recent success of encoder-decoder models for image segmentation to first identify traversable routes. The model performs supervised training  on a \textit{source domain} for traversability estimation, while domain generalization capabilities are achieved by pushing the model to learn representations shared by the \textit{source domain} and a \textit{target domain} (without annotations), employing an additional domain classifier and a gradient reversal layer that aims at generating indistinguishable representations between the two domains. This modification forces the traversability detection module to learn more general features, making the whole approach work even on scenarios for which labels are not available.

\begin{figure*}
    \centering
    \includegraphics[width=0.98\textwidth]{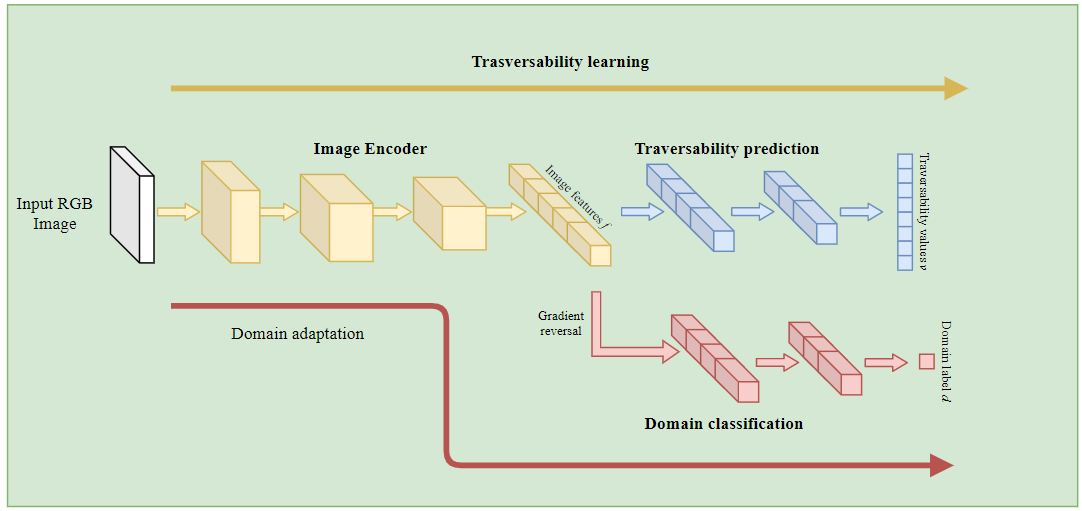}
    \caption{The overall architecture includes two main streams: a \emph{traversability learning} path that provides estimation on traversable regions in an input frame, and a \emph{domain adaptation} path that discriminates frames from the source and target domains. The gradient reversal layer enforces the image features $f$ learned by the image encoder to be domain-independent.}
    \label{fig:overall_architecture}
\end{figure*}

Fig.~\ref{fig:overall_architecture} shows the flowchart of the proposed method: at training time, it processes images with annotations from a source domain and images \textit{without} annotations from the target domain. Source domain annotations are used to train the model to identify route traversability (i.e., image encoder and traversability prediction streams), while unannotated target domain images are employed to enforce inter-domain representation similarity (i.e., training the domain classification branch). 

We pose our approach as a regression one, aiming at identifying  a set of real values, representing the maximum allowed distances along possible robot directions (see Fig.~\ref{fig:traversable}). Since it is important for the robot not to overcome the edges of traversability (because of the presence of obstacles or of visibility limits), we add to our model a \emph{safety-preserving loss}, whose main objective is to prevent potential collisions, by pushing the robot to favor early stopping than taking risks. 

Performance analysis made on a custom dataset, involving two different domains --- off-road (i.e., terrain) and on-road ---, shows that the proposed model achieves promising performance in both estimating accurately traversability paths and generalizing to unseen domains. 
To summarize, the main contributions of our paper are:
\begin{itemize}
    \item We propose a model able to analyze multiple traversability routes rather than only classifying frames as traversable/not traversable as in~\cite{HiroseSVGS18};
    \item Our model is able to generalize over different domains;
    \item Robot safety is taken into account, by training the model to make predictions that favor early stopping rather than risking collisions with obstacles.
\end{itemize}

%% file: related.tex
Traversability estimation is a core navigation module for robots interacting with dynamic and unstructured environments, as they must be able to move safely and avoid damages or injuries. 
While in indoor scenarios mobile robots are usually equipped with a map of the environments, which simplifies the traversability problem, in outdoor cases, it is rather challenging to define such maps in advance. Range sensors, such as lidars or depth cameras, may provide information on the environment morphology only, which, however, implies the need to define ad-hoc traversability metrics, thus lacking generalization. Hence, we focus on assessing traversability from RGB data. 

Traditionally, traversability estimation has been carried using hand-crafted features~\cite{10.5555/647288.721755,932633}. With the rediscovery of deep learning and its success in a variety of visual tasks, the paradigm has shifted to learning the information needed for the estimation directly from data.
Within the deep learning--based methods, the most common approach is to perform view synthesis of the scenes without obstacles~\cite{7139749,HiroseSVGS18,932633,HiroseSXMS19} and, at inference time, to measure the distance between acquired RGB frames and learned views. Thus, most of these methods are posed as anomaly detection tasks.
For example, the recent GoNET~\cite{HiroseSVGS18} model learns, using a generative adversarial network, to synthesize fish-eye images of traversable areas that are then compared, in pixel and feature spaces, to the current input frame. However, GoNET predicts traversability of the current view (right in front of the robot) at the current time step, limiting obstacle detection to the robot's immediate proximity. Our approach, instead, predicts traversability at a long temporal scale by estimating the traversable scores of different locations of the input frame. This strategy somehow recalls~\cite{HiroseSXMS19}, where multi-view synthesis is carried out to to estimate the traversability of multiple locations around the robot.
Although these methods have shown promising results, we argue that their limitation lies in the synthesis approach, i.e., learning distribution of scenes without obstacles greatly limits the generalization capabilities to unseen environments, possibly making them fail in cases of small deviations from the data used at training time. In other words, the model learns the features necessary to solve the training dataset, and cannot be employed beyond it.
We, instead, face the generalization problem by forcing the model to learn more general features that can be adopted in different scenarios. This is achieved by performing unsupervised domain adaptation at training time --- through the gradient reversal technique described in~\cite{ganin15} --- that prevents learned features from being strictly tied to the training domain.

%% file: method.tex
\subsection{Problem formulation}

We pose the traversability estimation task as a vector regression problem on image regions.

Let $\mathbf{I} \in \mathcal{I}$ be a tensor of size $C \times H \times W$ representing an RGB image and let $k$ be the number of same-width vertical areas into which the image can be divided: $\left\{ \mathbf{I}_1, \mathbf{I}_2, \dots, \mathbf{I}_k \right\}$, where:
\begin{equation}
\mathbf{I_i} = \mathbf{I}_{\left[:,:,\frac{iH}{k}:\frac{(i+1)H}{k}-1 \right]},
\end{equation}
each of size $C \times H \times \frac{W}{k}$ (assuming that $H$ is divisible by $k$, or suitably rounded otherwise), with subscripts expressed in Matlab-like notation. An example of the resulting subregions is shown in Fig.~\ref{fig:traversable}. Also, let $\mathbf{t}$ be a vector of size $k$, representing \emph{traversability scores} for each of the image subregions, ranging from 0 (not traversable) to 1 (fully traversable); details on how these values are assigned are given in Sect.~\ref{sec:dataset}.

The objective of the proposed method is to learn a function $f: \mathcal{I} \rightarrow [0,1]^k$ that, given an input image $\mathbf{I}$, provides an estimation $\mathbf{\tilde{t}} = f(\mathbf{I})$ that is as close as possible to the actual $\textbf{t}$ for a given image.

\subsection{Model architecture}

 The architecture of the employed model is based on a DeepLabV2~\cite{ChenPK0Y16} module for feature extraction, followed by a convolutional layer for dimensionality reduction and fully-connected layers providing the final $k$-dimensional output.

DeepLabV2 is a fully convolutional neural network, originally designed for semantic segmentation, that employs a ResNet-101~\cite{HeZRS15} backbone, atrous (or dilated) convolutions to reduce computational costs while keeping image resolution high, spatial pyramid pooling to enforce multi-scale analysis using a single-resolution input, and Conditional Random Field (CRF) post-processing to improve the final output maps. In our work, we employ DeepLabV2 as backbone of our architecture, with some simplifications aimed at keeping processing time low enough for real-time analysis (details in Sect.~\ref{sec:performance} for details): namely, we remove the multi-scale spatial pyramid analysis and the CRF post-processing, which becomes unnecessary as our output is not image-like.

The output of the DeepLabV2 encoder is a 2,048-channel feature map, which is reduced to 64 channels by a 1$\times$1 convolutional layer, including batch normalization on the input and a rectifier linear unit (ReLU) on the output. Due to the variable spatial size of the feature map, that depends on the size of the input image, at this stage we apply an adaptive pooling layer, in order to make the total number of neurons at this layer fixed. The resulting resized feature map is serialized and fed to a fully-connected layer, mapping to a $k$-dimensional vector representing the estimated $\mathbf{\tilde{t}}$ for input image $\mathbf{I}$. The architectural details for the the blocks following the DeepLabV2 backbone can be found in Tab.~\ref{tab:arch_details}.

\begin{table}[h]
\caption{Architectural details for the layers following the DeepLabV2 backbone in the proposed model. \newline Values $h$ and $w$ correspond to the spatial size of the feature maps returned by the backbone: for a 128 $\times$ 227 pixel input, $h$ and $w$ are 17 and 29, respectively.}
\label{tab:arch_details}
\begin{center}
\begin{tabular}{lcccc}
\toprule
\textbf{Layer}  & \textbf{Input size}           & \textbf{Kernel size} & \textbf{Output size} \\
\midrule
Batch norm.     & 2048$\times h \times w$       & -                    & 2048 $\times h \times w$ \\
Conv. 2D        & 2048$\times h \times w$       & 1 $\times$ 1         & 64 $\times h \times w$ \\
ReLU            & 64$\times h \times w$         & -                    & 64 $\times h \times w$ \\
Adaptive pool.  & 64$\times h \times w$         & -                    & 64 $\times$ 8 $\times$ 8 \\
Fully-connected & 4096                          & -                    & 9 \\
\bottomrule
\end{tabular}
\end{center}
\end{table}

\subsection{Safety-preserving loss}

Model training is carried out in a supervised way, by computing a loss function based on the difference between the estimated $\mathbf{\tilde{t}}$ and the correct $\mathbf{t}$ for a given image. In regression problems, the mean square error (MSE) loss is typically employed to make the model learn the correct predictions:

\begin{equation}
\mathcal{L} = \sum_{(\mathbf{I}, \mathbf{t})} \sum_{j=1}^k \left( t_{j} - \tilde{t}_j \right)^2 ,
\end{equation}
where $(\mathbf{I}, \mathbf{t})$ pair consists of an input image from a dataset and the corresponding target traversability values, and $\tilde{\mathbf{t}} = f(\mathbf{I})$.

However, the fact that the MSE loss is an even function may not be ideal to the task at hand. For example, let $t_i$ and $\tilde{t}_i$ be, respectively, the correct traversability value for a certain image subregion and the value predicted by the model. While the sign of $\left( t_i - \tilde{t}_i \right)$ does not affect the value of the MSE loss, it actually bears an important meaning on the safety of the prediction: if $t_i > \tilde{t}_i$, the model is predicting that a certain path is less traversable than it actually is, which may be a conservative yet safe prediction; however, if $t_i < \tilde{t}_i$, the model is overly and wrongly optimistic about the traversability of a certain path, which may lead to collisions with dangerous objects or overturnings, in the case of misjudged slopes of the surface.

Given these considerations, we modify the standard MSE loss to take into account a preference towards predicting ``safe'' traversability scores, i.e., such that $t_i > \tilde{t}_i$, by modifying the standard MSE loss as follows:

\begin{equation}
\mathcal{L}_s = \sum_{(\mathbf{I}, \mathbf{t})} \sum_{j=1}^k \left[ \left( t_{j} - \tilde{t}_j \right)^2 + \alpha \max\left(0, \tilde{t}_j - t_j \right)^2 \right] .
\end{equation}

The second term in the loss function, weighed by a hyperparameter $\alpha$, adds a further penalization to ``unsafe'' predictions, pushing the model to learn to be conservative, while still accurate --- thanks to the standard first term of the loss function.

In practice, we also include a regularization term in the loss function to reduce the risk of overfitting and instabilities, by minimizing the squared $L_2$ norm of the model's parameters, comprehensively indicated as a vector $\pmb{\theta}$:

\begin{equation}
\mathcal{L}_s = \sum_{(\mathbf{I}, \mathbf{t})} \sum_{j=1}^k \left[ \left(  \tilde{t}_j - t_{j} \right)^2 + \alpha \max\left(0, \tilde{t}_j - t_j \right)^2 \right] + \lambda \left\lVert \pmb{\theta} \right\lVert_2^2 ,
\end{equation}
where $\lambda$ controls the regularization strength.

The overall loss is minimized, during training, through gradient descent, i.e., by altering the parameters of the model towards the direction of local decrease of the loss function.

\subsection{Domain adaptation through gradient reversal}

Adaptation to different working conditions than seen in training is a major problem for deep learning models~\cite{ganin15,HausserFMC17,Mingsheng}, which are often able to replicate the performance achieved during training only when applied to input data that are close to the original distribution. However, the effort of adapting to a new environment may be high, mostly because of the need to collect and manually annotate new samples for model re-training. In our approach, we try to alleviate this problem by employing a \emph{domain adaptation} technique based on the use of a \emph{gradient reversal layer} during training.

In this scenario, we assume to have a \emph{source} dataset $\mathcal{D}_s$ of $(\mathbf{I}, \mathbf{t})$ pairs, consisting of images $\mathbf{I}$ with the corresponding traversability scores $\mathbf{t}$, and a \emph{target} dataset $\mathcal{D}_t$, whose samples include images only, \emph{with no traversability annotations}. In a practical situation, this setup would be equivalent to having a dataset with manual annotations on which the model is mainly trained, and a dataset corresponding to a new working environment, for which only image samples are available and no time or effort can be spent for generating annotations.

The objective of this new task is to train a model to predict correct traversability scores on the source dataset, while carrying out \emph{unsupervised domain adaptation} on the target dataset, i.e., training a model on annotated data while accounting for context changes in order to achieve comparable performance on the unannotated dataset. To this aim, we employ the \emph{gradient reversal} technique~\cite{ganin15}, illustrated in Fig.~\ref{fig:overall_architecture}. While the model for traversability prediction is trained on the source domain (since it requires annotations), we also train a \emph{domain classification} model on both the source and target domains to learn to discriminate whether an input image comes from one or the other. Part of the computation is shared between the two models: in particular, the input to the domain classifier is not directly an image, but an intermediate feature representation computed by the initial \emph{image encoding} portion of the traversability model. The key to performing domain adaptation is to reverse the gradients of the domain classification loss when backpropagating through the image encoder. This operation initiates a ``competition'' (similar in spirit to generative adversarial networks~\cite{NIPS2014_5423}), with the domain classifier attempting to correctly distinguish the two domains, and the image encoder trying to generate image features that are independent of the domain. As a result, the traversability prediction model tends to become independent of the domain of the original image, since the features it receives are domain-agnostic.

More formally, we first define the training set for the domain classifier as:

\begin{equation}
\mathcal{D}_d = \left\{ \left( \mathbf{I}, l_s \right) \right\}_{\left(\mathbf{I}, \mathbf{t} \right) \in \mathcal{D}_s} \cup \left\{ \left( \mathbf{I}, l_t \right) \right\}_{\mathbf{I} \in \mathcal{D}_t} ,
\end{equation}
where $l_s$ and $l_t$ are employed as \emph{domain labels} --- in practice, they are assigned the values 0 and 1.

Let $h(\mathbf{I})$ be the intermediate representation computed by the image encoder for input $\mathbf{I}$, and $g(h(\mathbf{I}))$ the output of the domain classifier, i.e., a scalar value estimating the likelihood of the input belonging to the target domain (with label $l_t = $1). The loss employed when training the domain classifier is the standard binary cross-entropy loss:

\begin{equation}
\mathcal{L}_d = - \sum_{\left( \mathbf{I}, l \right) \in \mathcal{D}_d} \Big[ l \log( g(h(\mathbf{I}))) + ( 1-l ) \log ( 1 - g(h(\mathbf{I})) ) \Big] ,
\end{equation}

with $l \in \{0,1\}$ being the domain label associated to input $\mathbf{I}$.

Architecturally, our domain classifier is a multi-layer perceptron, with an initial gradient reversal layer that simply changes the sign of the gradient of the domain classification loss when backpropagating along the image encoder, thus enforcing the above-described objective. As intermediate image features, we employ the output of the DeepLabV2 backbone of the traversability prediction model. At the output layer, a sigmoid activation is applied to scale the result as a probability value. The details of this domain classifier are presented in Tab.~\ref{tab:domain_arch_details}.

\begin{table}[h]
\caption{Architectural details for the domain classifier.}
\label{tab:domain_arch_details}
\begin{center}
\begin{tabular}{lcccc}
\toprule
\textbf{Layer}     & \textbf{Input size} & \textbf{Output size} \\
\midrule
Gradient reversal  & 4096 & 4096 \\
Fully-connected    & 4096 & 1024 \\
ReLU               & 1024 & 1024 \\
Fully-connected    & 1024 & 256 \\
ReLU               & 256  & 256 \\
Fully-connected    & 256  & 1 \\
Sigmoid            & 1    & 1 \\
\bottomrule
\end{tabular}
\end{center}
\end{table}

%% file: exp_intro.tex
This section describes the experiments we carried out to evaluate the
proposed methods in outdoor environments. 
We first test the capabilities of the whole framework in correctly assessing  traversability. We then evaluate the impact of our safety-preserving approach and model's domain generalization capabilities.
All code was written in Python using the PyTorch framework. Model's training was executed on an NVidia GeForce GTX TITAN X GPU, while on-field testing on the NVidia Jetson TX1 mounted on-board of the robot. 

%% file: dataset.tex
\begin{figure}
    \centering
    \includegraphics[width=0.47\textwidth]{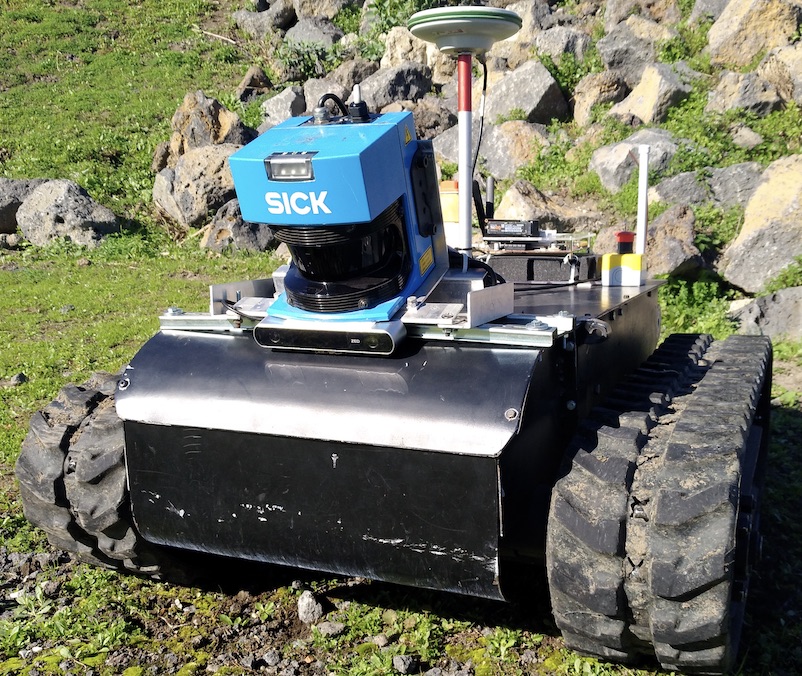}
    \caption{The tracked vehicle adopted during our experiments}
    \label{fig:track_vehicle}
\end{figure}

Experiments were carried out by teleoperating an unmanned ground robot employed for navigation in rough outdoor environments. The robot is a rubber-tracked vehicle (shown in Fig.~\ref{fig:track_vehicle}) through a maximum slope of 30\degree, and has an autonomy of 6-8 hours. Robot sensors include: motor encoders, an RTK-DGPS GNSS receiver, an attitude and heading reference system (AHRS), and a 2D lidar. An embedded acquisition and control system, based on the sbRIO-9626 board by National Instruments, acts as low-level interface with the sensors and actuators. 

A ZED stereo camera by Stereolabs placed in front of the robot acquires RGB 1280$\times$720 images at 15 fps. 
For our experiments, we used only right images from the stereo stream. 
The robot is also provided with two on-board companion computers: a Raspberry Pi3 B+, running the high-level control and navigation algorithms, and an NVidia Jetson TX1 (256 NVidia cores,  quad-core ARM A57 and 4 GB LPDDR4 memory) for traversability estimation inference.

\begin{figure*}
    \centering
    \includegraphics[width=1\textwidth]{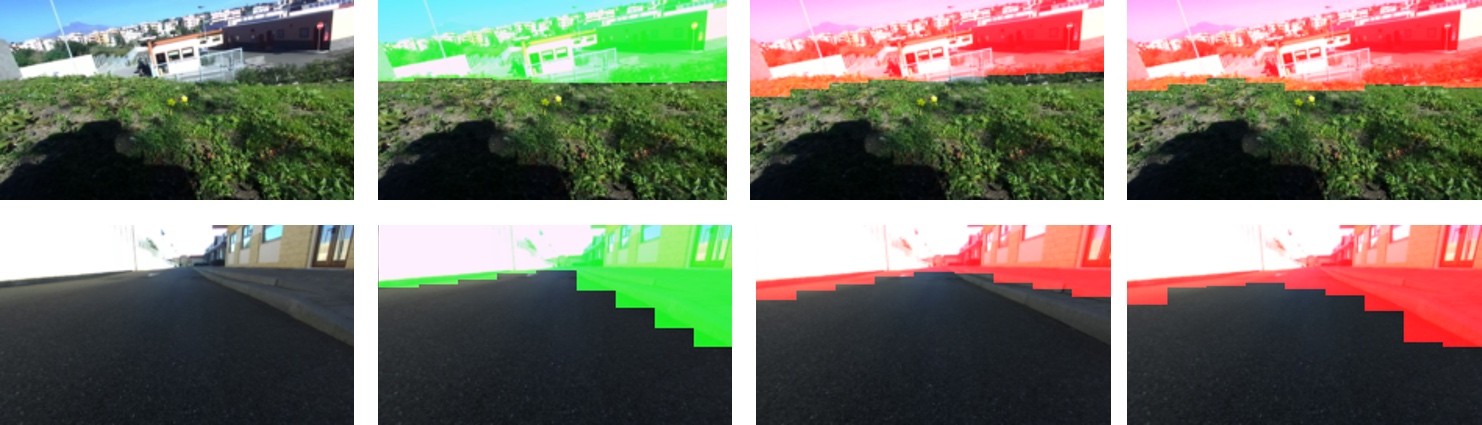}
    \caption{\textbf{Effect of safety-preserving loss}. From left to right: RGB frame, ground truth, model's estimation without safety-preserving loss, model's estimation with safety-preserving loss. Top row: off-road (terrain), bottom row: on-road. Safety-preserving loss makes estimation slightly less accurate, but more conservative.  }
    \label{fig:examples_full_no_safety}
\end{figure*}

RGB data, employed in our experiments, was acquired in two scenarios: \textit{on-road} and \textit{off-road} (terrain), see Fig.~\ref{fig:examples_full_no_safety} --- first column. In total, we collected 5,000 images (about 30 minutes of robot driving), equally split between the two scenarios. Images were combined with the position, in terms of Cartesian coordinates w.r.t. a local frame, and orientation of the robot. 
To ensure visual variability, we selected a subset of the 5,000 images, such that consecutive frames differ by a $\Theta_{th}$ yaw rotation or a $dist_{th}$  linear displacement of the robot, or a weighted combination of the two thresholds. This was necessary to remove near-duplicates which may affect training and test performance. 
In particular, the normalized angular difference $\Delta \theta$ is computed as:
\begin{equation}
    \Delta \theta = \frac{|\theta_i - \theta_j|}{\Theta_{th}}
\end{equation}
with $\theta_i$ being the yaw angle in the current frame, $\theta_j$ the yaw angle in the last selected frame, and $\Theta_{th}$ the threshold.
Since this difference is evaluated in terms of yaw angle, which takes values in the range [-180\degree,  180\degree], when the absolute value of difference exceeds 180\degree, it is subtracted to 360\degree.
The normalized linear displacement is evaluated in terms of Euclidean distance as:
\begin{equation}
    dist=\frac{\sqrt{(x_i-x_j)^2+(y_i-y_j)^2}}{dist_{th}}
\end{equation}
where $x_i$ and $y_i$ are the robot position coordinates in the current frame, $x_j$ and $y_j$ are the robot position coordinates in the last selected frame and $dist_{th}$ is the threshold.
Both values are normalized in order to be summed, i.e., $comb_{th}= dist + \Delta \theta$.

In our experiments, an image was considered meaningful  when $comb_{th} > 1$, and, in such a case, included in our dataset for model training and testing. 
Applying the above procedure, we selected 922 images out of the 5,000 recorded data (off-road: 419, on-road: 503), by setting $\Theta_{th} =$ 40\degree, because the horizontal field of view of the ZED camera is 85\degree, and $dist_{th} =$ 0.8 m, taking into account the maximum velocity of the considered robotic platform.

No post-processing was applied to the acquired data. Data annotation was carried out manually by a human operator, through an ad-hoc Matlab tool, who was asked to indicate the level of traversability for the robot, i.e., the level of trafficability of the robot until an obstacle is identified. More specifically, each image was divided into 9 vertical sections and for each section the human annotator had to draw in the image the line after which the robot could not go because of an obstacle or of loss of visibility (in Fig.~\ref{fig:examples_full_no_safety} --- second column --- non-traversable areas are shown in overlaid green bins). These annotations were then converted into numeric values (by normalizing the $y$ coordinate w.r.t. the image height) in the range $\left[0, 1\right]$, where 0 indicates non-traversable and 1 fully traversable.

%% file: training.tex
\subsubsection{Full dataset training}

In this setup, we train our traversability prediction model on both the ``off-road'' and ``on-road'' scenarios. This is the ideal training case, with annotations available for both domains.

We randomly split the dataset into a training set and a test set, with proportions 80\% (737 images) and 20\% (185 images), respectively. During training, images are resized so that the shortest side is 128 pixels: given the aspect ratio of the camera, the actual resolution of input images is 128$\times$227. Training is carried out for 200 epochs, using the Adam~\cite{KingmaB14} optimizer on mini-batches of size 16, with learning rate 2$\cdot$10$^{-4}$ and hyperparameters $\beta_1 =$ 0.5 and $\beta_2 =$ 0.999. The $\alpha$ parameter, weighing the safety-preserving loss, is set to 1.5. $L_2$ regularization is weighed by a factor $\lambda =$ 5$\cdot$10$^{-4}$. 

Test results are reported at the epoch for which the best training accuracy is achieved.

\subsubsection{Separate domains, without adaptation}

This setup serves as a baseline for our unsupervised domain adaptation approach, in order to assess the improvement obtained by integrating gradient reversal into the training procedure.

We separate the data from the two domains into a \emph{source} domain and a \emph{target} domain. For a fair comparison with the domain adaptation setup, we prepare the dataset splits for both setups in the same way, by splitting both the source domain and the target domain into a source training set, a source test set, a target training set and a target test set. Note that the exact sizes of those sets vary, depending on what domain is used as source and what as target. At this stage, the target training set is not employed; we only use the target test set as a benchmark to assess performance when switching to a different domain.

The same parameters for training as described above are employed. Test results are reported at the epoch for which the best training accuracy (on the source training set) is achieved.

\subsubsection{Separate domains, with adaptation}

In this setup, we integrate gradient reversal into the training procedure. In detail, we employ the target training set to train the traversability prediction model, and both the target and test training sets to train the domain classifier and apply gradient reversal. The training procedure for the traversability prediction model is as described above. In order to train the domain classifier, we additionally employ an SGD optimizer, with learning rate 0.001 and momentum 0.9.

As above, test results are reported at the epoch when the best training accuracy on the source training set is achieved.

%% file: performance.tex
\subsubsection{Supervised traversability training}

In this experiment, we evaluate the accuracy of the traversability prediction model, trained with supervision, in the following cases: a) using the full dataset, with no domain separation, b) using data from the ``off-road'' domain only; c) using data from the ``on-road'' domain only. 
The accuracy achieved in this setup represents an upper bound of the proposed method, when trained on both the ``off-road'' and ``on-road'' scenarios, with all annotations available. 
As baseline, we use a prediction model based on ResNet-101~\cite{HeZRS15} (a state-of-the-art network for visual classification), suitably modified for traversability estimation. In particular, we remove the last classification layer and append the same additional layers described in Tab.~\ref{tab:arch_details}. We start from ResNet-101 model pre-trained on ImageNet, and fine-tune the baseline model on our training data.
Since also our model, in the encoder part, already includes a ResNet-101 backbone for semantic segmentation, the baseline helps to understand the role of: a) manipulating the learned embedding (i.e., the decoder part of segmentation model of DeepLabV2) and, b) employing multiscale analysis through atrous convolutions. 

\begin{figure*}
    \centering
    \includegraphics[width=0.24\textwidth]{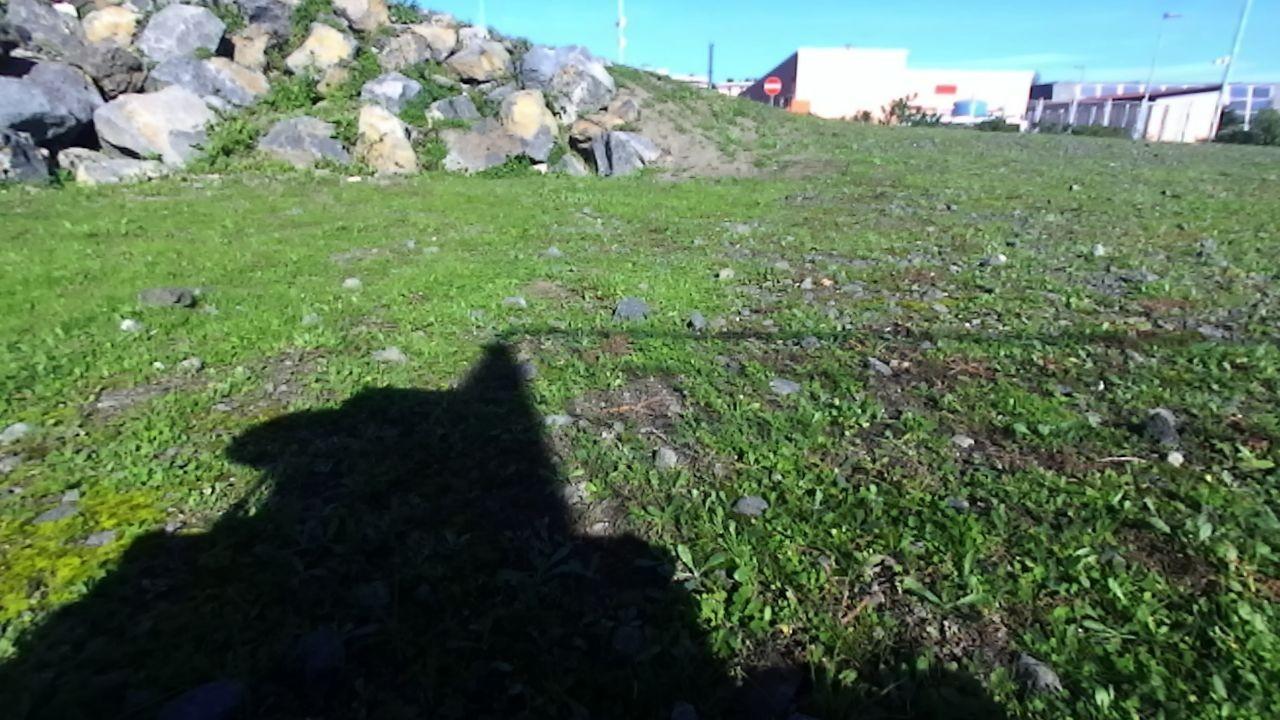}\hspace{0.1cm}
    \includegraphics[width=0.24\textwidth]{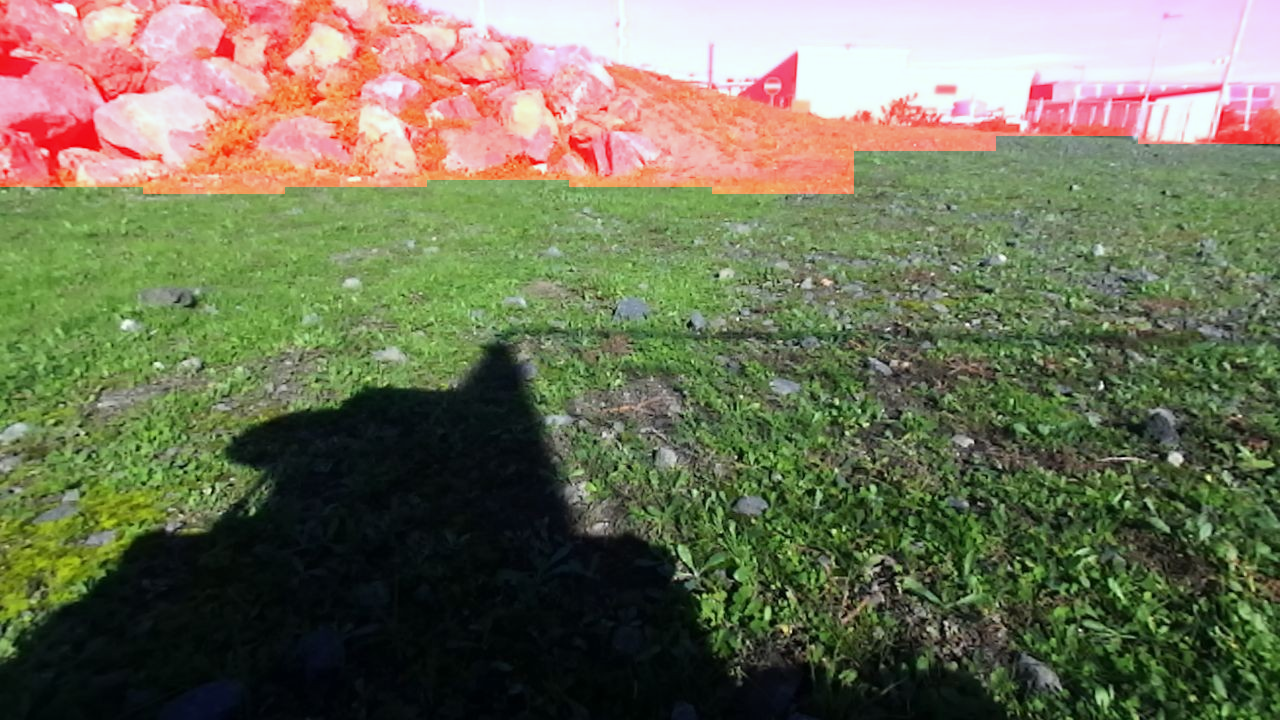}\hspace{0.1cm}
    \includegraphics[width=0.24\textwidth]{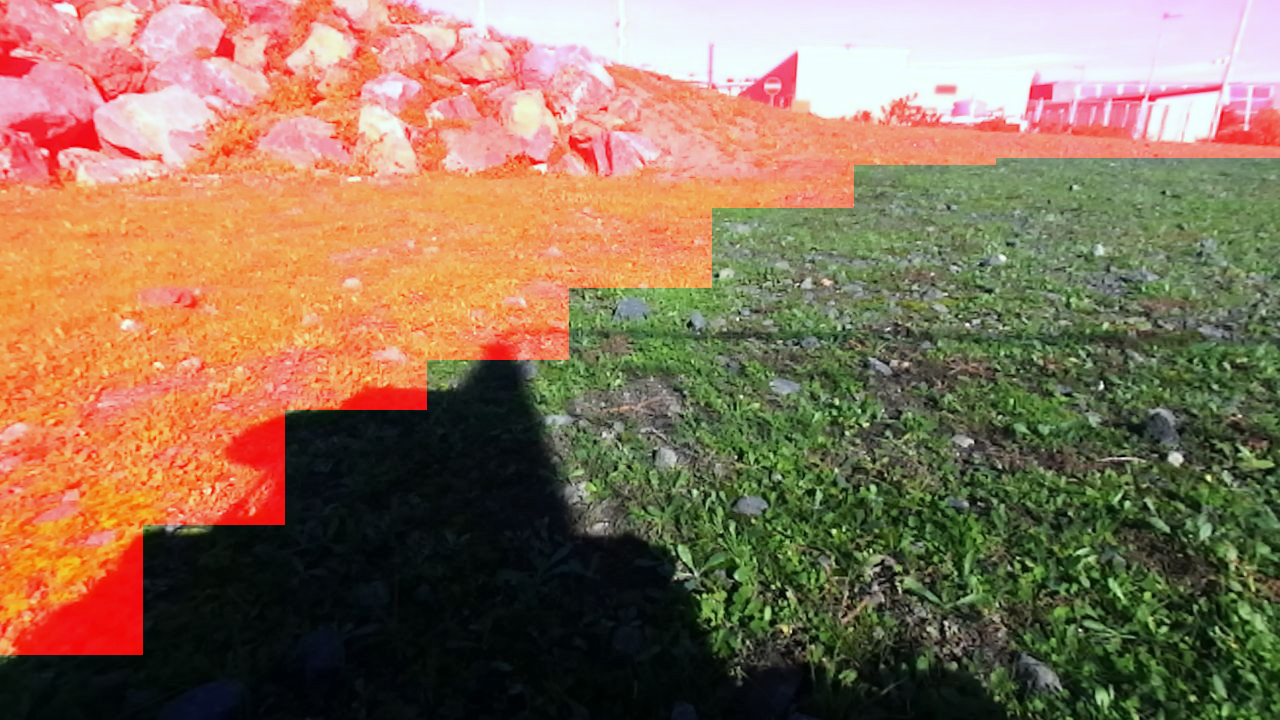}\hspace{0.1cm}
    \includegraphics[width=0.24\textwidth]{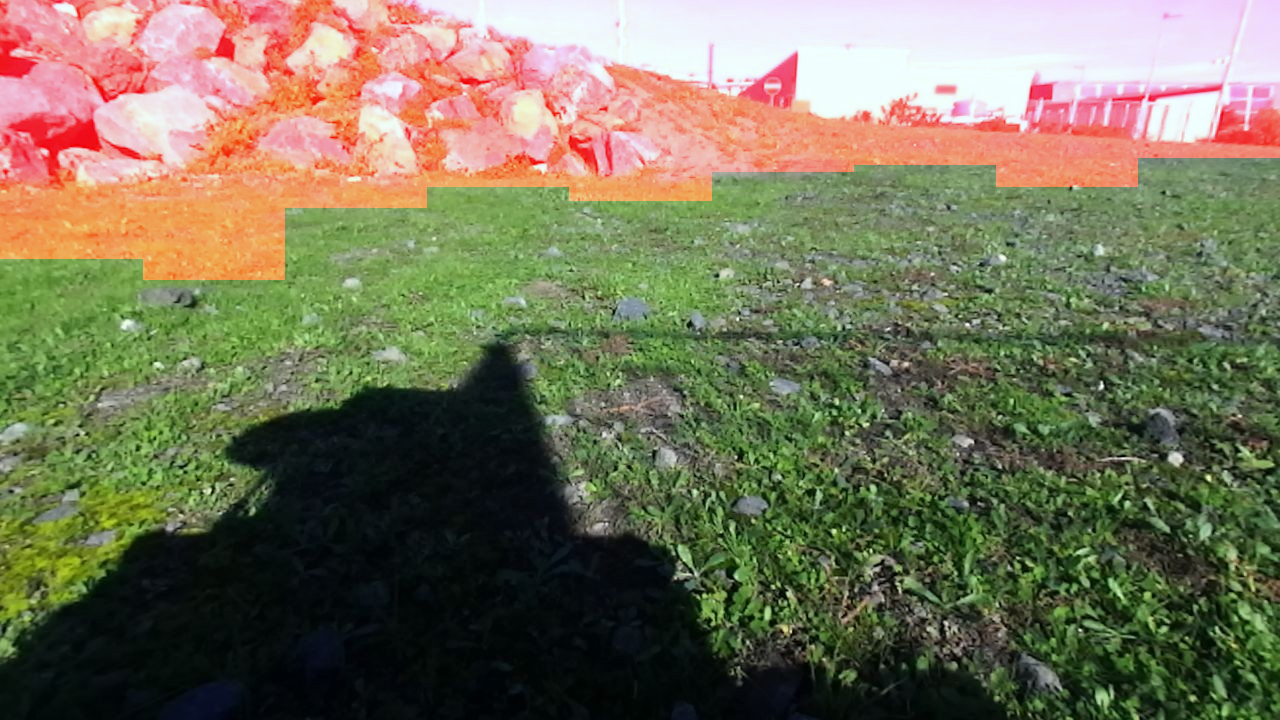}
    \caption{\textbf{Domain adaptation}. From left to right: input frame from ``off-road'' domain; model prediction when trained the same domain;  model prediction when trained on ``on-road'' domain, without adaptation; model prediction when trained on ``on-road'' domain, with unsupervised adaptation on ``off-road''.}
    \label{fig:domain_adaption}
\end{figure*}

\begin{table}[htb!]
\caption{Test accuracy on the supervised learning setup}\vspace{-0.1cm}
\label{tab:results}
\begin{center}
\begin{tabular}{cccc}
\toprule
\textbf{Model} & \textbf{MAE$_\text{all}$} & \textbf{MAE$_\text{on\_road}$} & \textbf{MAE$_\text{off\_road}$}\\
\midrule
ResNet-101    & 0.091 & 0.095 & 0.141 \\
Our approach    & 0.049 & 0.031 & 0.062\\
Safety-preserving loss &  0.084 & 0.075 & 0.110\\
\bottomrule
\end{tabular}
\end{center}
\end{table}

Results, in terms of mean absolute error (MAE) between the ground-truth traversability score and the model's estimation, are presented in Tab.~\ref{tab:results}, where we also compare the accuracy achieved when enabling or disabling our safety-preserving loss. 
It is possible to notice that: a) employing a semantic-segmentation backbone (as in our approach) yields better performance than using a standard classifier backbone (e.g., ResNet-101); this means that, for reliable traversability assessment, it is necessary to build a detailed representation of the scene (as obtained through atrous convolutions and decoder layers used in our approach) rather than a holistic one as when using a classifier backbone only; b) the variant without safety-preserving loss achieves a slightly higher accuracy, which can be explained by the fact that the model is less constrained on the features it is allowed to learn; indeed, the safety-preserving loss forces the model to satisfy both the overall constraint on mean square error \emph{and} the additional constraint on the sign of the error. However, the difference can be better appreciated from a qualitative point of view. Fig.~\ref{fig:examples_full_no_safety} shows some examples of the model's outputs on test images, without and with the safety-preserving loss, respectively. In the figures, colored bars are overlaid to the untraversable portions of the image: green bars denote ground-truth annotations, and red bars denote model outputs. While the output of the model trained without safety-preserving loss is quite accurate  (third column of Fig.~\ref{fig:examples_full_no_safety}), it is possible to notice that in some cases the traversability prediction is too optimistic, compared to the actual annotation. Fig.~\ref{fig:examples_full_no_safety} (fourth column), instead, shows that, at the cost of a slight reduction in accuracy, the predictions obtained when enabling the safety-preserving loss are more conservative and reliable for actual navigation.

\subsubsection{Separate domains, without adaptation}

Next, we evaluate the accuracy of the traversability prediction model when training on one domain and testing on the other. The results, unsurprisingly lower than the previous experiment, are also shown in Tab.~\ref{tab:results_adapt}. This experiment is meant to provide a lower bound on the accuracy of the proposed domain adaptation method.

\subsubsection{Separate domains, with adaptation}

Finally, we evaluate the impact that unsupervised domain adaptation through gradient reversal has, when training on one domain and testing on the other. The comparison in Tab.~\ref{tab:results_adapt} shows that, although our domain adaptation approach achieves --- predictably --- a lower accuracy than supervised training (see accuracy in Tab.~\ref{tab:results}), the improvement on the setup without adaptation is significant. 
\begin{table}[h!]
\caption{Domain adaptation accuracy on different training scenarios.}\vspace{-0.5cm}
\label{tab:results_adapt}
\begin{center}
\begin{tabular}{lcccc}
\toprule
\textbf{Training} & \textbf{Test} & \textbf{Safety-preserving} & \textbf{Grad-Reversal} & \textbf{MAE} \\
\midrule
\multicolumn{5}{l}{\textit{Separate domains, no adaptation}}\\
\midrule
Off-Road & On-Road    & \checkmark & --  & 0.449\\
On-Road    & Off-Road & \checkmark & --  & 0.413\\
\midrule
\multicolumn{5}{l}{\textit{Separate domains, with adaptation}}\\
\midrule
Off-Road & On-Road    & \checkmark & \checkmark & 0.291\\
On-Road    & Off-Road & \checkmark & \checkmark & 0.249\\
\bottomrule
\end{tabular}
\end{center}
\end{table}

Fig.~\ref{fig:domain_adaption} shows qualitatively the performance achieved by our traversability prediction model when training on a domain and testing on the other, with and without our unsupervised adaptation approach. In the example, an input frame (first figure) from the ``off-road'' (i.e., terrain) domain is used as test sample in three different setups: same-domain training and testing (second figure); training on ``on-road'' and testing on ``off-road'', without adaptation (third figure); and training on ``on-road'' and testing on ``off-road'', with adaptation (fourth figure). While the best accuracy is achieved, as expected, when performing same-domain training and testing, it is interesting to notice how prediction changes without and with adaptation. In the third figure, the model seems confused as to whether grass (in the off-road images) is traversable or not: it marks it as not traversable in correspondence to its own shadow (which perhaps may resemble on-road asphalt) and as traversable when no vertical reference is provided. However, when domain adaptation is performed, the model seems more confident about the traversability of grass, with a slight uncertainty (always on the safe side) on the top region of the image.

These results confirm the effectiveness of domain adaptation techniques in a real-world use case, opening a promising line of research towards improving the generalizability of data-driven mobile robot navigation approaches.

Processing times for model's inference are of 120 ms per frame on the on-board Jetson TX1 GPU, thus allowing the traversability estimation model to run at about 8 frames per second. GPU memory occupation is of 760 MB, out of the total 4 GB available on the Jetson TX1.
The domain adaptation approach, through  gradient reversal, does not affect inference processing times as it is executed only during model training. 
\subsection{Robot Navigation}

Although robot navigation is beyond the scope of this paper, we also evaluate the qualitative behavior of the robot when employing the developed traversability inference model. 
The traversability values predicted by our deep model are translated into linear and angular velocity commands, in order to drive the robot towards the most traversable section of the incoming image. 
In particular, the predicted traversability score of the central section of the image controls robot linear velocity: if it is over 0.5 (i.e., half of the front section is traversable), velocity is set to the maximum allowed value, otherwise linear velocity is reduced until 0 when traversing score is equal (or lower than) 0.1 (i.e., there is not enough traversable area in front of the robot). 
The angular speed is, instead, set to point towards the area with highest traversing score.
On-field tests reveal promising results, as the robot is capable to steer, decelerate or stop, in order to avoid obstacles, cliffs or steep paths.

%% file: conclusions.tex
In this paper we propose an approach for traversability estimation on mobile robots that addresses two important issues: a) safety preservation, as security and safety of robots and people is a most pressing need, especially in human-robot interaction scenarios, b) generalizability of the prediction, in the line of research toward pushing robot performance closer to human adaptation capabilities, without requiring new annotation efforts for re-training on new scenarios.
More specifically, we have demonstrated that our approach is able to perform well in a scenario different than the training one, without any additional supervision. Moreover, posing the traversability estimation task as a prediction of future possible actions can facilitate joint traversability and navigation policy learning, for example, by making the model predicting also steering angles as in~\cite{8264734}.

As a future work, we aim to address another key aspect of mobile robots, the robustness of vision algorithms against adversarial attacks. Indeed,  although deep learning models are showing impressive performance in several vision tasks, they can be fooled by adversarial examples~\cite{Dezfooli16}, i.e., images altered by a small perturbations, suitably designed to mislead the output prediction. This poses serious security issues especially in cases of traversability estimation of mobile robots interacting with people or in high-risk environments, as simple visual distractors may lead to unexpected and potentially-dangerous results.

%% file: root.bbl
\begin{thebibliography}{10}
\providecommand{\url}[1]{#1}
\csname url@rmstyle\endcsname
\providecommand{\newblock}{\relax}
\providecommand{\bibinfo}[2]{#2}
\providecommand\BIBentrySTDinterwordspacing{\spaceskip=0pt\relax}
\providecommand\BIBentryALTinterwordstretchfactor{4}
\providecommand\BIBentryALTinterwordspacing{\spaceskip=\fontdimen2\font plus
\BIBentryALTinterwordstretchfactor\fontdimen3\font minus
  \fontdimen4\font\relax}
\providecommand\BIBforeignlanguage[2]{{%
\expandafter\ifx\csname l@#1\endcsname\relax
\typeout{** WARNING: IEEEtran.bst: No hyphenation pattern has been}%
\typeout{** loaded for the language `#1'. Using the pattern for}%
\typeout{** the default language instead.}%
\else
\language=\csname l@#1\endcsname
\fi
#2}}

\bibitem{6696421}
N.~{Hirose}, R.~{Tajima}, and K.~{Sukigara}, ``Personal robot assisting
  transportation to support active human life — posture stabilization based
  on feedback compensation of lateral acceleration,'' in \emph{2013 IEEE/RSJ
  International Conference on Intelligent Robots and Systems}, Nov 2013, pp.
  659--664.

\bibitem{2832747.2832901}
M.~Veloso, J.~Biswas, B.~Coltin, and S.~Rosenthal, ``Cobots: Robust symbiotic
  autonomous mobile service robots,'' in \emph{Proceedings of the 24th
  International Conference on Artificial Intelligence}, ser. IJCAI’15.\hskip
  1em plus 0.5em minus 0.4em\relax AAAI Press, 2015, p. 4423–4429.

\bibitem{5675357}
N.~{Hirose}, K.~{Sukigara}, H.~{Kajima}, and M.~{Yamaoka}, ``Mode switching
  control for a personal mobility robot based on initial value compensation,''
  in \emph{IECON 2010 - 36th Annual Conference on IEEE Industrial Electronics
  Society}, Nov 2010, pp. 1914--1919.

\bibitem{6719358}
B.~{Cafaro}, M.~{Gianni}, F.~{Pirri}, M.~{Ruiz}, and A.~{Sinha}, ``Terrain
  traversability in rescue environments,'' in \emph{2013 IEEE International
  Symposium on Safety, Security, and Rescue Robotics (SSRR)}, Oct 2013, pp.
  1--8.

\bibitem{narvaez2018terrain}
F.~Y. Narv{\'a}ez, E.~Gregorio, A.~Escol{\`a}, J.~R. Rosell-Polo,
  M.~Torres-Torriti, and F.~A. Cheein, ``Terrain classification using {T}o{F}
  sensors for the enhancement of agricultural machinery traversability,''
  \emph{Journal of Terramechanics}, vol.~76, pp. 1--13, 2018.

\bibitem{doi:10.1002/rob.21833}
\BIBentryALTinterwordspacing
K.~Skonieczny, D.~K. Shukla, M.~Faragalli, M.~Cole, and K.~D. Iagnemma,
  ``Data-driven mobility risk prediction for planetary rovers,'' \emph{Journal
  of Field Robotics}, vol.~36, no.~2, pp. 475--491, 2019. [Online]. Available:
  \url{https://onlinelibrary.wiley.com/doi/abs/10.1002/rob.21833}
\BIBentrySTDinterwordspacing

\bibitem{7989182}
M.~{Pfeiffer}, M.~{Schaeuble}, J.~{Nieto}, R.~{Siegwart}, and C.~{Cadena},
  ``From perception to decision: A data-driven approach to end-to-end motion
  planning for autonomous ground robots,'' in \emph{2017 IEEE International
  Conference on Robotics and Automation (ICRA)}, May 2017, pp. 1527--1533.

\bibitem{7139749}
B.~{Suger}, B.~{Steder}, and W.~{Burgard}, ``Traversability analysis for mobile
  robots in outdoor environments: A semi-supervised learning approach based on
  3{D}-lidar data,'' in \emph{2015 IEEE International Conference on Robotics
  and Automation (ICRA)}, May 2015, pp. 3941--3946.

\bibitem{6698836}
I.~{Bogoslavskyi}, O.~{Vysotska}, J.~{Serafin}, G.~{Grisetti}, and
  C.~{Stachniss}, ``Efficient traversability analysis for mobile robots using
  the kinect sensor,'' in \emph{2013 European Conference on Mobile Robots},
  Sep. 2013, pp. 158--163.

\bibitem{balta2013terrain}
H.~Balta, G.~De~Cubber, D.~Doroftei, Y.~Baudoin, and H.~Sahli, ``Terrain
  traversability analysis for off-road robots using time-of-flight 3d
  sensing,'' in \emph{7th IARP International Workshop on Robotics for Risky
  Environment-Extreme Robotics, Saint-Petersburg, Russia}, 2013.

\bibitem{973332}
H.~{Koyasu}, J.~{Miura}, and Y.~{Shirai}, ``Real-time omnidirectional stereo
  for obstacle detection and tracking in dynamic environments,'' in
  \emph{Proceedings 2001 IEEE/RSJ International Conference on Intelligent
  Robots and Systems. Expanding the Societal Role of Robotics in the the Next
  Millennium (Cat. No.01CH37180)}, vol.~1, Oct 2001, pp. 31--36 vol.1.

\bibitem{4399610}
{D. Kim}, {S. Oh}, and J.~M. {Rehg}, ``Traversability classification for {UGV}
  navigation: a comparison of patch and superpixel representations,'' in
  \emph{2007 IEEE/RSJ International Conference on Intelligent Robots and
  Systems}, Oct 2007, pp. 3166--3173.

\bibitem{10.5555/647288.721755}
I.~Ulrich and I.~R. Nourbakhsh, ``Appearance-based obstacle detection with
  monocular color vision,'' in \emph{17th National Conference on Artificial
  Intelligence and Innovative Applications of Artificial Intelligence}.\hskip
  1em plus 0.5em minus 0.4em\relax AAAI Press, 2000, p. 866–871.

\bibitem{HiroseSVGS18}
N.~Hirose, A.~Sadeghian, M.~V{\'{a}}zquez, P.~Goebel, and S.~Savarese, ``Gonet:
  {A} semi-supervised deep learning approach for traversability estimation,''
  in \emph{{IROS} 2018}.\hskip 1em plus 0.5em minus 0.4em\relax {IEEE}, 2018,
  pp. 3044--3051.

\bibitem{932633}
P.~H. {Batavia} and S.~{Singh}, ``Obstacle detection using adaptive color
  segmentation and color stereo homography,'' in \emph{IEEE International
  Conference on Robotics and Automation 2001}, vol.~1, May 2001, pp. 705--710
  vol.1.

\bibitem{HiroseSXMS19}
N.~Hirose, A.~Sadeghian, F.~Xia, R.~Mart{\'{\i}}n{-}Mart{\'{\i}}n, and
  S.~Savarese, ``Vunet: Dynamic scene view synthesis for traversability
  estimation using an {RGB} camera,'' \emph{{IEEE} Robotics and Automation
  Letters}, vol.~4, no.~2, pp. 2062--2069, 2019.

\bibitem{ganin15}
Y.~Ganin and V.~Lempitsky, ``Unsupervised domain adaptation by
  backpropagation,'' in \emph{Proceedings of the 32nd International Conference
  on Machine Learning}, ser. Machine Learning Research, F.~Bach and D.~Blei,
  Eds., vol.~37, 2015, pp. 1180--1189.

\bibitem{ChenPK0Y16}
L.~Chen, G.~Papandreou, I.~Kokkinos, K.~Murphy, and A.~L. Yuille, ``Deeplab:
  Semantic image segmentation with deep convolutional nets, atrous convolution,
  and fully connected crfs,'' \emph{CoRR}, vol. abs/1606.00915, 2016.

\bibitem{HeZRS15}
K.~He, X.~Zhang, S.~Ren, and J.~Sun, ``Deep residual learning for image
  recognition,'' \emph{CoRR}, vol. abs/1512.03385, 2015.

\bibitem{HausserFMC17}
P.~H{\"{a}}usser, T.~Frerix, A.~Mordvintsev, and D.~Cremers, ``Associative
  domain adaptation,'' in \emph{ICCV 2017}.\hskip 1em plus 0.5em minus
  0.4em\relax {IEEE} Computer Society, 2017, pp. 2784--2792.

\bibitem{Mingsheng}
M.~Long, Y.~Cao, J.~Wang, and M.~I. Jordan, ``Learning transferable features
  with deep adaptation networks,'' in \emph{ICML 2018}.\hskip 1em plus 0.5em
  minus 0.4em\relax JMLR.org, 2015, p. 97–105.

\bibitem{NIPS2014_5423}
I.~Goodfellow, J.~Pouget-Abadie, M.~Mirza, B.~Xu, D.~Warde-Farley, S.~Ozair,
  A.~Courville, and Y.~Bengio, ``Generative adversarial nets,'' in \emph{NIPS
  2014}, 2014.

\bibitem{KingmaB14}
D.~P. Kingma and J.~Ba, ``Adam: {A} method for stochastic optimization,'' in
  \emph{ICLR 2015}, Y.~Bengio and Y.~LeCun, Eds., 2014.

\bibitem{8264734}
A.~{Loquercio}, A.~I. {Maqueda}, C.~R. {del-Blanco}, and D.~{Scaramuzza},
  ``Dronet: Learning to fly by driving,'' \emph{IEEE Robotics and Automation
  Letters}, vol.~3, no.~2, pp. 1088--1095, April 2018.

\bibitem{Dezfooli16}
S.~Moosavi{-}Dezfooli, A.~Fawzi, and P.~Frossard, ``Deepfool: {A} simple and
  accurate method to fool deep neural networks,'' in \emph{2016 {IEEE}
  Conference on Computer Vision and Pattern Recognition, {CVPR} 2016, Las
  Vegas, NV, USA, June 27-30, 2016}.\hskip 1em plus 0.5em minus 0.4em\relax
  {IEEE} Computer Society, 2016, pp. 2574--2582.

\end{thebibliography}
